\documentclass[10pt, a4paper]{article}

\usepackage{lrec2026} 

\usepackage{graphicx}
\usepackage{todonotes}
\usepackage{xcolor}
\usepackage{xspace}
\usepackage{float}
\usepackage{multirow}
\usepackage{amsmath}

\usepackage{xspace}
\newcommand{\elrl}{\textsc{ELRL}\xspace}
\newcommand{\elrls}{\textsc{ELRLs}\xspace}

\newcommand{\hrl}{\textsc{HRL}\xspace}

\title{Evaluating Extremely Low-Resource Machine Translation: A Comparative Study of ChrF++ and BLEU Metrics}

\name{Sanjeev Kumar, Preethi Jyothi, Pushpak Bhattacharyya} 

\address{Dept. of CSE, IIT Bombay, \\
         Mumbai, India \\
         \{sanjeev, pjyothi\}@cse.iitb.ac.in\\}

\abstract{
Evaluating machine translation (MT) quality in extremely low-resource language (\elrl) scenarios poses unique challenges, as widely used metrics such as BLEU, effective in high-resource settings, often misrepresent quality in data-scarce contexts. This work presents a comparative analysis of BLEU, an n-gram-based metric, and ChrF++, a character-based metric, for MT evaluation in \elrl settings. We examine how each metric responds to translation artifacts, including hallucinations, repetition, source-text copying, and diacritic (\textit{matra}) variations across three \elrls: Magahi, Bhojpuri, and Chhattisgarhi, with a focus on outputs from large language models (LLMs) and neural MT (NMT) systems. While recent work often relies solely on ChrF++, our findings show that BLEU, despite its lower absolute scores, provides complementary lexical-precision insights that improve interpretability. 
 \\ \newline  }

\begin{document}

\maketitleabstract

\section{Introduction}
MT has made remarkable progress in recent years, driven by large parallel corpora and advancements in language models. However, the scarcity of data in most languages makes MT development and evaluation difficult in \elrl scenarios. Traditional MT evaluation metrics like BLEU \cite{papineni-etal-2002-bleu}, widely adopted and effective for high-resource languages (HRL), often struggle to reflect translation quality in data-scarce settings accurately. Since BLEU relies on n-gram overlap, it is highly sensitive to word order and tends to \emph{penalize morphologically rich languages}.

The emergence of Large Language Models (LLMs) \cite{brown2020language, chen2021evaluating, ouyang2022training, hadi2023large, achiam2023gpt} complicate evaluation in low-resource (LR) settings. These models, while powerful, are prone to errors such as hallucinations, repetitive outputs, and source copying, especially when translating from a related \hrl to an \elrl \cite{guerreiro2023hallucinations}. Character-based metrics like ChrF++ \cite{popovic-2017-chrf} often fail to detect these issues, rewarding surface overlap. BLEU, conversely, can over-penalize legitimate morphological or word-order variation, yielding low scores.

On the other hand, with smaller architectures and limited training data, neural MT systems often produce shorter or incomplete outputs. BLEU’s strict n-gram precision and brevity penalty yield disproportionately low scores, reflecting brevity and reordering sensitivity. In Indic \elrls, diacritics (\textit{matras}) and morphological inflection further exaggerate these penalties, unlike character-based metrics such as ChrF++. Prior work has observed similar issues in morphologically rich or low-data languages \cite{maillard-etal-2023-small, hus-anastasopoulos-2024-back, lu-etal-2025-low, tran-etal-2024-irish}. In contrast, ChrF++ can yield higher scores for the same outputs, especially when scripts overlap. This divergence highlights blind spots when only one metric is used. Therefore, the reliability of BLEU and ChrF++ in \elrl scenarios, remains uncertain. Recent studies \cite{tanzer2023benchmark, iyer-etal-2024-quality, wu-etal-2024-far, lippmann-etal-2025-context} rely primarily on ChrF++, whereas we argue that both metrics together yield a more complete assessment.

While BLEU–ChrF++ divergences are not unique to low-resource languages, but their impact intensifies under extreme data scarcity. Limited parallel data causes lexical sparsity, leading models to over-copy or under-generate, which magnifies metric differences. Shared scripts among Indo-Aryan languages further inflate ChrF++ via surface overlap, even when semantic fidelity is poor. In \hrl, richer training coverage and standardized orthography mitigate these distortions, allowing BLEU and ChrF++ to correlate more closely. Hence, the same linguistic phenomena display more severely and more unpredictably under the \elrl setting. These effects are further magnified in multilingual models trained unevenly across languages.

Given these limitations, learned evaluation metrics such as COMET \cite{rei-etal-2020-comet} and BLEURT \cite{sellam-etal-2020-bleurt} appear to be good alternatives. However, their reliability in out-of-domain languages such as \elrls remains uncertain. For closely related languages, they may implicitly map system outputs to an \hrl seen during training (e.g., treating Magahi as Hindi), inflating scores despite semantic divergence. Such biases underscore the need for caution when applying learned metrics to \elrl evaluation and motivate our focus on a systematic investigation of BLEU and ChrF++ divergences in both LLM and NMT-based systems. We recommend practitioners jointly inspect BLEU and ChrF++ scores for \elrl MT, as their divergence can signal specific linguistic issues such as hallucination, source copying, and orthographic errors. We further provide case-by-case interpretations of these divergences to guide more informed metric use in \elrl evaluation. Our contributions are:
\begin{enumerate}
    \item  We compare BLEU and ChrF++ metrics for MT in \elrls, including their combined use and variations.
    \item Evaluation of LLM-based (Aya-101, Airavata) and NMT-based (mT5-Large) model between English and four Indo-Aryan languages: Hindi, Magahi, Bhojpuri, and Chhattisgarhi (with the last three being ELRLs).
    \item We analyze the linguistically grounded interplay between ChrF++ and BLEU, examining scenarios where both metrics align or diverge, leveraging linguistic insights to interpret their variations.
\end{enumerate}

\section{Background}
BLEU remains the most widely used MT evaluation metric due to its simplicity and correlation with human judgments. However, it is highly sensitive to word order and lexical variation, often penalizing valid translations that differ morphologically or syntactically from the reference. Other word-based metrics, such as METEOR \cite{banerjee-lavie-2005-meteor} and TER \cite{snover-etal-2006-study}, attempt to address these issues but still struggle in LR contexts. To address this, character-based metrics such as ChrF \cite{popovic-2017-chrf} were proposed, offering better robustness to inflectional variation and subword overlap.

In \elrls, with minor word or character variations, causes significant errors. BLEU struggles to capture such errors, while LLMs in \elrl settings introduce issues like hallucinations \cite{guerreiro2023hallucinations}, repetition, and source copying, further complicating evaluation. In Indic \elrl, diacritics (\textit{matras}) impact accuracy, penalizing BLEU, whereas character-based metrics like ChrF++ are less sensitive to such variations. The reliability of BLEU and ChrF++ for \elrl evaluation, especially using LLMs, is uncertain. In Indic languages, ChrF++ may inflate scores by favoring source copying, while BLEU can be overly harsh due to limited lexical overlap despite meaning preservation. 


\citet{kocmi-etal-2024-navigating} investigates the relationship between metric magnitudes and perceived translation quality across 94 languages included in BERT \cite{devlin-etal-2019-bert} or XLM-RoBERTa \cite{conneau2019unsupervised} training; their analysis spans a broad set of metrics, including learned ones such as COMET and BLEURT. In contrast, our study focuses exclusively on \elrl settings and undertakes a detailed qualitative examination of BLEU and ChrF++ in both LLM and NMT-based systems. We specifically explore how variations in these metrics reflect translation artifacts such as hallucinations, source copying, and repetition, which are common in \elrl outputs. Unlike \citet{kocmi-etal-2024-navigating}, whose primary goal is to correlate multiple metrics with human judgments to establish quality thresholds, our work delivers \elrl-specific, linguistically grounded insights into what BLEU–ChrF++ divergences reveal about translation quality and underlying linguistic phenomena.

While human evaluation could address many of these limitations, it poses significant logistical barriers in \elrl contexts: a shortage of qualified annotators, geographically dispersed speaker communities, and the absence of standardized annotation protocols. Even when feasible, human evaluation is resource-intensive and difficult to replicate. This motivates our focus on automated metrics. By systematically comparing BLEU and ChrF++ for MT evaluation in Indic \elrls, we analyze how their interplay reflects translation artifacts such as hallucinations, repetition, source copying, and morphological variation. By doing so, we provide \elrl-specific, linguistically grounded insights into metric behavior and offer case-to-case interpretation of such metrics in \elrl settings.

\section{Experiments and Results}
\subsection{Experimental Setup} \label{exp-details}
We evaluate Aya-101 \cite{ustun-etal-2024-aya}, mT5-Large \cite{xue-etal-2021-mt5}, and Airavata \cite{gala2024airavata} on translation tasks for Bhojpuri, Chhattisgarhi, and Magahi in both English/Hindi→Target and reverse directions. Aya-101 is a translation-focused multilingual model (101 languages), while Airavata is trained exclusively on Indic languages with a higher proportion of Hindi data, making both well-suited for Indo-Aryan \elrls. To cover different model families, mT5-Large is included as a representative neural MT model. All models are fine-tuned on the 6,192-sentence NLLB Seed corpus \cite{maillard-etal-2023-small} and evaluated on the 1,012-sentence FLORES-200 devtest set \cite{goyal-etal-2022-flores}. BLEU and ChrF++ are computed using the SacreBLEU \cite{post-2018-call} with standard tokenization. To observe diverse translation behaviors beyond deterministic decoding, we enable stochastic generation (\texttt{do\_sample=True}) during inference, allowing multiple plausible hypotheses per input. We fine-tune using PEFT (LoRA; \cite{hu2022lora}) with rank 16 and scaling 32, applied to query/value projections, keeping the pretrained weights frozen.

\begin{table}[]
    \centering
    \small
    \begin{tabular}{|c|c|c|c|} \hline
         \textbf{Direction} & \textbf{ChrF}++ & \textbf{BLEU} & \textbf{Model Name} \\ \hline
         hi $\rightarrow$ bho & 27.69 & 6.76 & Aya-101 \\
         hi $\rightarrow$ bho & 27.62 & 11.35 & mT5-Large \\ \hline
         en $\rightarrow$ bho & 20.15 & 7.57  & Airavata \\
         en $\rightarrow$ bho & 21.53 & 11.55 & mT5-base\\ \hline
         en $\rightarrow$ bho & 23.82 & 19.69  & mT5-Large \\
         hi $\rightarrow$ bho & 22.37 & 7.01 & mT5-Large$^{2}$\\ \hline
         en $\rightarrow$ bho & 15.04 & 0.96  & Airavata$^{2}$ \\
         en $\rightarrow$ bho & 15.39 & 2.15 & mT5-Large$^{2}$\\ \hline
    \end{tabular}
    \caption{BLEU–ChrF++ results for English/Hindi $\to$ Bhojpuri. $^{2}$ Trained on 20\% less data.}
    \label{tab:to_bhojpuri}
\end{table}
\subsection{Results} \label{exp-result}
Tables~\ref{tab:to_bhojpuri}, \ref{tab:for_magahi}, and \ref{tab:to_chhattisgarhi} present results for English and Hindi translations into Bhojpuri, Magahi, and Chhattisgarhi, respectively. Each table is categorized by target language and translation direction (Hindi/English $\leftrightarrow$ Target), with each block representing a unique model–language pair. Thus, Tables~1,~2,~and~5 each summarize one target language, ensuring that comparisons are made only within, not across, language pairs.  

We observe large disparities between BLEU and ChrF++ scores: even small ChrF++ shifts often correspond to major BLEU changes across models. While ChrF++ is popular for low-resource translation, it can overestimate quality for closely related languages that share scripts. For example, in Hindi$\rightarrow$Magahi, source copying yields a high ChrF++ (41.43) but a low BLEU (18.09) (Table~\ref{tab:for_magahi}). Conversely, a significant BLEU rise with only a modest ChrF++ gain indicates genuine quality improvement. These divergences underscore the need to use multiple metrics for reliable \elrl MT evaluation.

\begin{table}[]
    \centering
    \small
    \begin{tabular}{|c|c|c|c|} \hline
        \textbf{Direction} & \textbf{ChrF}++ & \textbf{BLEU} & \textbf{Model Name} \\ \hline
        en  $\rightarrow$ mag & 27.26 & 9.81 & Airavata \\
        en  $\rightarrow$ mag & 26.63 & 12.54 & MT5-large \\ \hline
        hi  $\rightarrow$ mag & 41.43 & 18.09 & MT5-large \\
        hi  $\rightarrow$ mag & 39.9 & 11 & MT5-large$^{2}$  \\ \hline
        en  $\rightarrow$ mag & 43.26 & 35.77 & Aya-101 \\
        hi  $\rightarrow$ mag & 41.43 & 18.09 & MT5-large \\ \hline
        \multicolumn{4}{c}{} \\ \hline
        mag $\rightarrow$ hi & 43.86 & 39.44 & Aya-101 \\
        mag $\rightarrow$ hi & 44.78 & 29.36 & MT5-large \\ \hline
        mag $\rightarrow$ en & 42.28 & 20.11 & MT5-large \\
        mag $\rightarrow$ en & 43.57 & 14.83 & MT5-large$^{2}$  \\ \hline
    \end{tabular}
    
    \caption{BLEU–ChrF++ results for target language (English/Hindi $\leftrightarrow$ Target). $^{2}$ Trained on 20\% less data.}
    \label{tab:for_magahi}
\end{table}

\begin{table}[ht]
    \centering
    \small
    \begin{tabular}{|c|c|c|c|} \hline
        \textbf{Direction} & \textbf{ChrF++} & \textbf{BLEU} & \textbf{Model Name} \\ \hline
        en $\rightarrow$ hne & 18.46 & 7.93 & Airavata \\
        hi $\rightarrow$ hne & 17.55 & 15.14 & mT5-Large \\ 
        hi $\rightarrow$ hne & 19.47 & 7.66 & Aya-101 \\ \hline
        en $\rightarrow$ hne & 23.83 & 19.59 & mT5-Large$^{2}$ \\
        en $\rightarrow$ hne & 24.34 & 29.55 & mT5-Large \\ \hline
        en $\rightarrow$ hne & 18.28 & 3.15 & Airavata$^{2}$ \\
        hi $\rightarrow$ hne & 17.55 & 15.14 & mT5-Large \\ \hline
    \end{tabular}

    \caption{BLEU–ChrF++ results for target language (English/Hindi $\leftrightarrow$ Target). $^{2}$ Trained on 20\% less data.}
    \label{tab:to_chhattisgarhi}
\end{table}

\subsection{Analysis} \label{analysis}
\begin{table*}[h!]
\centering
\scriptsize
\setlength{\tabcolsep}{2pt}
\begin{tabular}{|p{2cm}|p{6cm}|p{5cm}|c|c|}
\hline
\multicolumn{1}{|c|}{\textbf{Case}} & \multicolumn{1}{|c|}{\textbf{Original (English)}} & \multicolumn{1}{|c|}{\textbf{Output}} & \textbf{BLEU} & \textbf{ChrF++} \\ \hline

\multirow{2}{=}{Decrease in both ChrF++ and BLEU} & Antim sanskar mein shamil hoe ke lel char million se adhik logan Rome gel rahalei.  & \textbf{Output 1.} Char million log maut ke dauraan Roma galay. & 5.1 & 19.96 \\ \cline{3-5} 

 & \textit{(More than four million people had gone to Rome to attend the funeral.)} & \textbf{Output 2.} Amrityu ke dekhkar paanch million log Roma gelei. & 2.74 & 15.61 \\ \hline
 
\multirow{2}{=}{Stable ChrF++ with significant change in BLEU} & U ego prasuti rog visheshagya ke roop me prashikshan lalakay aur 1959 me Auckland ke rashtriya mahila aspatal me kaam kare le shuru kailkay.  & \textbf{Output 1.} U stanpanavid ke roop me shikshit holkhin aur 1959 me Aaskel ke rashtriya mahila aspatal me kary kare lagi shuru holkhin. & 23.12 & 45.92 \\ \cline{3-5} 

 & \textit{(He trained as an obstetrician and began working at Auckland’s National Women’s Hospital in 1959.)} & \textbf{Output 2.} Ek rogi ke roop me shikshit holai aur 1959 me Aulakelet ke rashtriya mahila aspatal me kaam karne shuru kalai. & 12.63 & 44.94 \\ \hline
 
\multirow{2}{2cm}{Increase in ChrF++ and decrease in BLEU} & Pichhle mahinE ego Rashtrapati Aayog desh ke naye chunaav ke or le jaay ke upaay ke package ke tahat poorv CEP ke isteefe ke sifarish kailkay hal.  & \textbf{Output 1.} Kai mahinE pehle ek sansadiya commission desh ke naya chunaav tak pahunche ke lel poorv CEP ke padasthapan ke prastav delkay & 7.4 & 27.3 \\ \cline{3-5}

 & \textit{(Last month, a presidential commission recommended the resignation of the former CEP as part of a package of measures to lead the country toward new elections.)} & \textbf{Output 2.} Hal hi mein Rashtrapati Samiti poorv EP ke narazgi ke desh mein naye chunaavon ke disha mein ek pack-up mein sathe upyog kare ke lel sifarish kailkay. & 2.33 & 30.53 \\ \hline
 
\multirow{2}{2cm}{Increase in ChrF++ and minor change in BLEU} & NateejaTan, do machhli prajatiyan vilupt ho galay hal, aaur do anya luptapraya ho galay hay, jekara Humpback Chub bhi shamil hay. & \textbf{Output 1.} Phalswaroop, do paudha prajati khatre hue hay, aaur dunu anya atyadhik khatre hue hay, jahan hi Mopbki Chub bhi hay & 5.29 & 20.48 \\ \cline{3-5}

 & \textit{(As a result, two fish species became extinct, and two others became endangered, including the Humpback Chub.)} & \textbf{Output 2.} NateejaTan, duno paudha prajati gayab ho gelei, aaur do anya sankraman mein pahunch gelei, jonme Humpback Chub shamil hai. & 5.3 & 27.59 \\ \hline
 
\multirow{2}{2cm}{Slight increase in ChrF++ and significant increase in BLEU} & U din mein aaspaas ke satah se thande aaur raat mein garm hovo hay.  & \textbf{Output 1.} U din mein aaspaas ke satah tulna mein sajni hai aaur raat mein garm hai. & 30.62 & 55.35 \\ \cline{3-5}
 & \textit{(During the day, it is cooler than the surrounding surface, and at night, it becomes warmer.)} & \textbf{Output 2.} U din mein aaspaas ke satah se behtar hay aur raat mein kam garm hay. & 35.83 & 57.49 \\ \hline
 
\multirow{2}{2cm}{ChrF++ decreases and BLEU increases} & Chhutti manabe ke lel antardeshiy jalmaarg ego accha vishay ho sakai chhalei.  & \textbf{Output 1.} Ego samudri jal maarg ek tyohar par aadharit kare ke lel ego accha tattva ho sakai hai & 7.92 & 35.83 \\ \cline{3-5}

 & \textit{(Inland waterways could be a great option for vacationing.)} & \textbf{Output 2.} Paryavaraniy pani ke nadiyon ke ego accha vishay ban sakai hai. & 14.47 & 28.95 \\ \hline
 
\end{tabular}
\caption{Comparison of translation outputs across representative cases, 
with English translations provided in parentheses. Examples illustrate six 
typical BLEU–ChrF++ divergence patterns. Scores are computed on the original script; transliterated outputs are shown only for readability.}
\label{tab:example-table}
\end{table*}
We examine variations of BLEU and ChrF++, offering deeper insights into translation quality that help detect issues like hallucinations, repetition, duplication, and source copying. The following six settings illustrate these patterns. While we illustrate each error type with a single representative example for clarity, these are drawn from a larger set of outputs exhibiting the same pattern. The aggregated metric behaviors for all such cases are reflected in Tables~\ref{tab:to_bhojpuri}, ~\ref{tab:for_magahi}, and ~\ref{tab:to_chhattisgarhi}, ensuring that our observations are not based on isolated instances but on consistent trends across the dataset. Representative cases shown here were selected from the diverse hypotheses generated during inference to illustrate each recurring pattern observed across the FLORES-200 devtest set. We define ``stable'' as $\Delta< \pm 1$, ``minor change'' as $1 \leq |\Delta| \leq 3$, and ``significant'' as $|\Delta| > 3$ for both ChrF++ and BLEU.


\noindent \textbf{1. Decrease in Both ChrF++ and BLEU.} Indicates poor translation quality or wrong-script output. For instance, in Table \ref{tab:example-table} in English$\rightarrow$Bhojpuri, the model produces partial English, dropping both metrics due to meaning distortion and structural mismatch. BLEU penalizes n-gram misalignment; ChrF++ drops from reduced character overlap.

\noindent \textbf{2. Stable ChrF++ with Significant Changes in BLEU.} Hallucinated outputs cause a sharp BLEU to drop sharply while ChrF++ remains steady, reflecting character-level overlap despite lexical divergence. In Table \ref{tab:example-table}, ChrF++ shifts marginally from 45.92 to 44.92, while BLEU halves (23.1-to-12.6), showing that surface similarity can mask poor adequacy.

\noindent \textbf{3. Increase in ChrF++, Decrease in BLEU.} Typical of partial source copying. For Magahi$\to$Hindi (Table \ref{tab:for_magahi}) using Aya-101 and mT5-large, the output remains largely in Hindi. While ChrF++ slightly increases (43.86 to 44.78) due to character overlap, the BLEU score drops significantly (from 39.44 to 29.36), indicating poor n-gram alignment and translation quality. Copying is not always harmful in closely related languages, where lexical overlap can yield comprehensible results, but it reduces lexical diversity captured by BLEU.

\noindent \textbf{4. Increase in ChrF++ with Minor Change in BLEU.} This pattern indicates out-of-context word generation. While ChrF++ increases due to character overlap (e.g., hi$\rightarrow$bho and en$\rightarrow$bho in Table \ref{tab:to_bhojpuri}, from 20.15 to 27.69), BLEU drops (7.57 to 6.76), showing that higher character-level matches do not always ensure better translations. At the sentence level (Table \ref{tab:example-table}), Output 1 misinterprets \textit{machhli prajatiyan} (fish species) as \textit{paudha prajati} (plant species), raising ChrF++ slightly despite semantic distortion, lowering BLEU.

\noindent \textbf{5. Minor Increase in ChrF++ with a Significant Increase in BLEU.} 
This pattern reflects improved lexical formation, often from accurate diacritics (\textit{matras}) and better word alignment. In Table~\ref{tab:for_magahi}, for English$\rightarrow$Magahi and Hindi$\rightarrow$Magahi, ChrF++ rises moderately (41.43$\rightarrow$43.26) while BLEU nearly doubles (18.09$\rightarrow$35.77), indicating stronger n-gram precision and fluency. In Table~\ref{tab:example-table}, Output 1 modifies \textit{thande} (cold) to \textit{tulna mein sajni hai} (comparatively cool); Output 2 refines it to \textit{behtar} (better) and \textit{kam garm} (less warm), producing a more natural translation. BLEU rewards improved n-gram matches, whereas ChrF++ changes little due to minor surface edits.

\noindent \textbf{6. ChrF++ Decreases, and BLEU Increases.} 
Here, character overlap drops but lexical precision improves through longer, more accurate n-grams a frequent pattern in morphologically rich languages. BLEU rises with stronger word- and phrase-level matches, while ChrF++ declines as alternative lexical forms reduce character similarity. For English$\rightarrow$Magahi (Table~\ref{tab:for_magahi}), ChrF++ falls (39.9$\rightarrow$26.63) yet BLEU increases (11$\rightarrow$12.54), reflecting better structure despite surface divergence. In Table~\ref{tab:example-table}, Output 1 replaces \textit{antardeshiy jalmaarg} (inland waterways) with \textit{samudri jal maarg} (marine waterways), and Output 2 generalizes to \textit{paryavaraniy pani ke nadiyan} (environmental water bodies), yielding a more natural but less literal translation. 
Borderline cases (e.g., 2 vs 5, 3 vs 6) are distinguished by thresholding $\Delta$ to ensure mutually exclusive definitions.


\section{Conclusion}
This study examined MT evaluation challenges for \elrls, focusing on LLMs and NMT models. Our findings highlight the limitations of using only ChrF++ for \elrls without examining BLEU.
By evaluating different models and translation directions, we show how linguistic proximity, morphology, and data scarcity affect these metrics. Our results suggest that a single metric may not be sufficient for evaluating \elrl translations and that a combination of metrics is necessary for a more reliable \elrl evaluation.
We recommend practitioners jointly inspect BLEU and ChrF++ scores for \elrl MT, as their divergence can signal specific linguistic issues such as hallucination, source copying, and orthographic errors.
This combined interpretation yields more robust and interpretable evaluations than relying on either metric alone.

\section{Limitations}
Our study focuses on only three Indic \elrls in translation with English and Hindi, which may not fully capture the challenges and variations present in other low-resource languages. The findings may not generalize to \elrls with different linguistic structures, scripts, or typological characteristics.

\section{Ethics Statement}
All experiments were conducted on publicly available datasets, including the FLORES-200 \cite{goyal-etal-2022-flores} and NLLB Seed corpus \cite{maillard-etal-2023-small}. No private, user-generated, or sensitive data were used.   
Our study focuses on evaluation methodology, and no human participants were involved. We acknowledge potential biases inherent in pre-trained multilingual models such as Aya-101 and mT5, particularly regarding underrepresented languages.

\section{Bibliographical References}\label{sec:reference}

\bibliographystyle{lrec2026-natbib}
\bibliography{custom}


\appendix

\end{document}